\documentclass[conference]{IEEEtran}
\pdfoutput=1 
\IEEEoverridecommandlockouts
\usepackage{cite}
\usepackage{amsmath,amssymb,amsfonts}
\usepackage{graphicx}
\usepackage{textcomp}
\usepackage{multirow}
\usepackage[table,xcdraw]{xcolor}
\usepackage{xcolor}
\usepackage{algpseudocode}
\usepackage{algorithm}

\def\BibTeX{{\rm B\kern-.05em{\sc i\kern-.025em b}\kern-.08em
    T\kern-.1667em\lower.7ex\hbox{E}\kern-.125emX}}
    
\begin{document}

\title{OCT-SelfNet: A Self-Supervised Framework with Multi-Modal Datasets for Generalized and Robust Retinal Disease Detection \\

}
\author{\IEEEauthorblockN{Fatema-E- Jannat, Sina Gholami, Minhaj Nur Alam, Hamed Tabkhi}
\IEEEauthorblockA{\textit{Dept. of Electrical and Computer Engineering} \\
\textit{University of North Carolina at Charlotte}\\
Charlotte, North Carolina, USA\\
\{fjannat, sgholami, minhaj.alam, htabkhiv\}@charlotte.edu}

}

\maketitle

\begin{abstract}
Despite the revolutionary impact of AI and the development of locally trained algorithms, achieving widespread generalized learning from multi-modal data in medical AI remains a significant challenge. This gap hinders the practical deployment of scalable medical AI solutions. Addressing this challenge, our research contributes a self-supervised robust machine learning framework, OCT-SelfNet, for detecting eye diseases using optical coherence tomography (OCT) images. In this work, various data sets from various institutions are combined enabling a more comprehensive range of representation. Our method addresses the issue using a two-phase training approach that combines self-supervised pretraining and supervised fine-tuning with a mask autoencoder based on the SwinV2 backbone by providing a solution for real-world clinical deployment. Extensive experiments on three datasets with different encoder backbones, low data settings, unseen data settings, and the effect of augmentation show that our method outperforms the baseline model, Resnet-50 by consistently attaining AUC-ROC performance surpassing 77\% across all tests, whereas the baseline model exceeds 54\%. Moreover, in terms of the AUC-PR metric, our proposed method exceeded 42\%, showcasing a substantial increase of at least 10\% in performance compared to the baseline, which exceeded only 33\%. This contributes to our understanding of our approach’s potential and emphasizes its usefulness in clinical settings
\end{abstract}

\begin{IEEEkeywords}
AI, Self-supervised, Supervised, Transformer, SwinV2, ViT, MAE, Autoencoder, OCT, classification, Multi-modal
\end{IEEEkeywords}

\section{Introduction}

Within the realm of medical artificial intelligence (AI), there is a significant gap between the progress made in AI technology resulting in the development of various locally trained AI algorithms and the challenge of using these models on a varied patient population on a large scale. This limits the practical deployment of scalable medical AI solutions in real-world scenarios. In our research, we have addressed two main issues. Firstly, robust machine learning (ML) tools can detect eye diseases using optical coherence tomography (OCT) images in managing the eye. Secondly, we addressed the challenge of creating a model that can learn generalized features from diverse unlabeled data to make it widely applicable in real-world scenarios.

Age-related macular degeneration (AMD) is one of the leading causes of irreversible blindness and vision impairment (VI) globally. VI affects nearly 2.2 billion people globally, among which almost 1 billion cases could be prevented with early diagnosis and intervention \cite{who_blindness_visual_impairment}. Therefore, it is essential to identify individuals at risk of disease onset or progression from early to more advanced stages since timely intervention can prevent or slow progression, thus preventing irreversible VI \cite{scott2013long}. Individuals with a high risk for VI would benefit from more frequent ophthalmic examinations, monitoring, and treatment \cite{yi2009spectral}. 

ML and deep learning (DL) techniques have demonstrated significant promise in automated ophthalmic diagnosis \cite{friberg2011analysis, alam2020quantitative, schmidt2018prediction, wang2016genetic}. However, using similar datasets during training often hinders their real-world effectiveness, resulting in challenges when deployed in clinical settings. To optimize their effectiveness in clinical workflow, they require access to diverse datasets from multiple institutions with varying demographics, OCT image-capturing devices, or protocols to improve their adaptability, versatility, and scalability. 

This work is one of the first to bring and evaluate recent advances in large pre-trained transformers to enhance the detection and diagnosis in biomedical imaging, focusing on automated ophthalmic diagnosis.  We have developed a large-scale, self-supervised model with random masking inspired by masked autoencoder \cite{he2022masked} with SwinV2\cite{liu2022swin} backbone, explicitly for classifying AMD using OCT images. More specifically, we focus on the binary classification of distinguishing normal cases from those with AMD. By leveraging self-supervised learning (SSL), this model aims to reduce the need for extensive manual annotations by experts, easing the workload and facilitating broader clinical AI applications from retinal imaging data. Our model can learn versatile and generalizable features from unlabeled retinal OCT datasets, which is crucial for creating AI systems that require fewer labeled examples to adapt to various diagnostic tasks. 

In contrast to previous works \cite{leandro2023oct, tsuji2020classification, LEE2017322, 8120661, lu2018deep} which studies one dataset in isolation, our study is distinct and intrinsically more challenging as we examine the domain adaptation complexity by considering multiple datasets. Multiple OCT datasets, Kermany et al.\cite{kermany}, Srinivasan et al.\cite{srinivasan}, Li et al.\cite{li2020octa}, named as DS1, DS2, and DS3 respectively, were combined to train a robust DL model to increase the training data's diversity and allow our model to learn a broader range of patterns and relationships. This diversity will improve the model's generalization to new, unseen data, and it will be beneficial for the cases where the larger dataset is unavailable. The model is trained and validated with the training and validation set of DS1, DS2, and DS3 during the self-supervised pre-training stage. It is subsequently fine-tuned independently on each dataset and cross-evaluated on all test sets. The obtained results are then compared with the baseline model, Resnet-50, which is trained on each dataset individually and cross-evaluated across all test sets. Our proposed framework, OCT-SelfNet, consistently demonstrates better performance, frequently matching or exceeding the baseline model's AUC-ROC and AUC-PR values. Notably, our OCT-SelfNet-SwinV2 model, fine-tuned on DS1, demonstrated remarkable AUC-ROC performance, scoring 0.96, 0.99, and 0.93 on Test Set-1, Test Set-2, and Test Set-3, respectively. In contrast, the baseline model achieved scores of 0.98, 0.99, and 0.56 on the same test sets, emphasizing the better discriminative abilities of our model. This trend persisted across other datasets as well, as seen in DS2, where our method exceeded the baseline with AUC-ROC scores of 0.79, 0.99, and 0.86, compared to the baseline's scores of 0.59, 0.80, and 0.54. Furthermore, on DS3, our approach yielded AUC-ROC scores of 0.75, 0.93, and 0.99, surpassing the baseline scores of 0.72, 0.87, and 0.96.  We even assess the model's performance under reduced labeled training data, specifically down to 50 percent, to evaluate its robustness in handling small labeled datasets.

These consistent improvements showcase the robust performance of our proposed methodology across diverse datasets. This similar trend is observed in the case of the AUC-PR metric as well, further underscoring the consistency of our OCT-SelfNet-SwinV2 model over the baseline across different datasets.


The main contributions of this paper are summarized as follows:

\begin{itemize}
    \item This paper is one of the first to bring large self-supervised pre-training vision transformers to biomedical imaging, focusing on classifying AMD using OCT images and demonstrating the benefits of domain generalization across multiple publicly available datasets.
    
    \item  This paper introduces a two-phase approach: (1) SwinV2-based masked autoencoder in the pre-training phase to understand better the image structure and relationships between different regions and (2) fine-tuning stage classifier for learning the specific AMD using OCT use cases and classification tasks.
    
    \item Through a comprehensive evaluation and ablation study, this paper demonstrates that the proposed approach shows much higher robustness and better generalization even when evaluated on different test sets without requiring additional fine-tuning. This is promising for effective implementation in practical clinical environments.
\end{itemize}

\begin{figure*}[h!]
\centerline{\includegraphics[width=.98\linewidth]{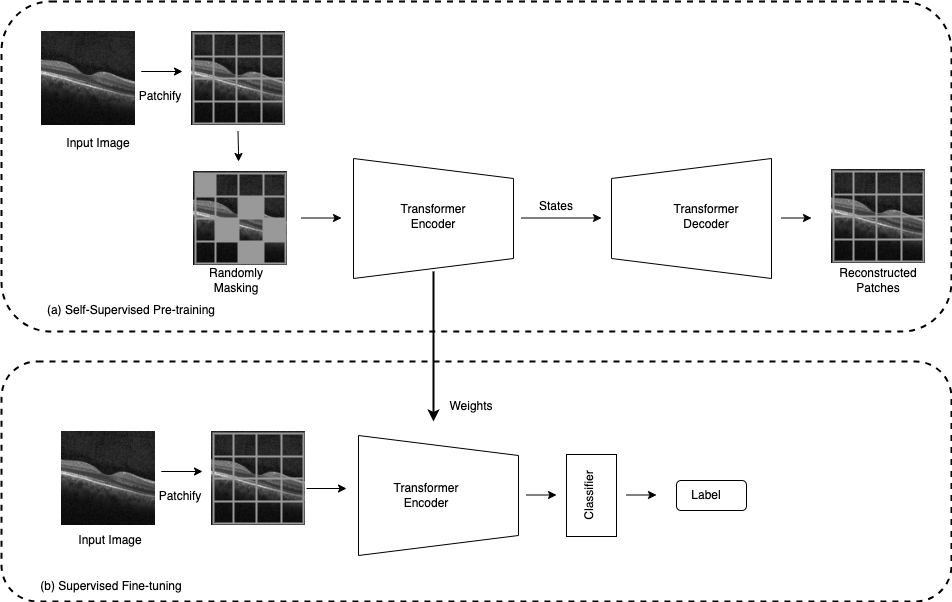}}
\caption{Overview of the framework: In the initial pre-training phase (upper section), the framework utilizes masked image autoencoder as a self-supervised task to learn representations from unlabeled images. In this process, a random subset of image patches is masked and fed into the auto-encoder. Then in the subsequent fine-tuning stage (lower section), the pre-trained encoder from the first phase is employed along with a linear classifier for the classification task. The learned weights from the pre-training phase are transferred to the fine-tuning phase.}
\label{fig}
\end{figure*}

\begin{figure*}[h!t]
\centerline{\includegraphics[width=1\linewidth]{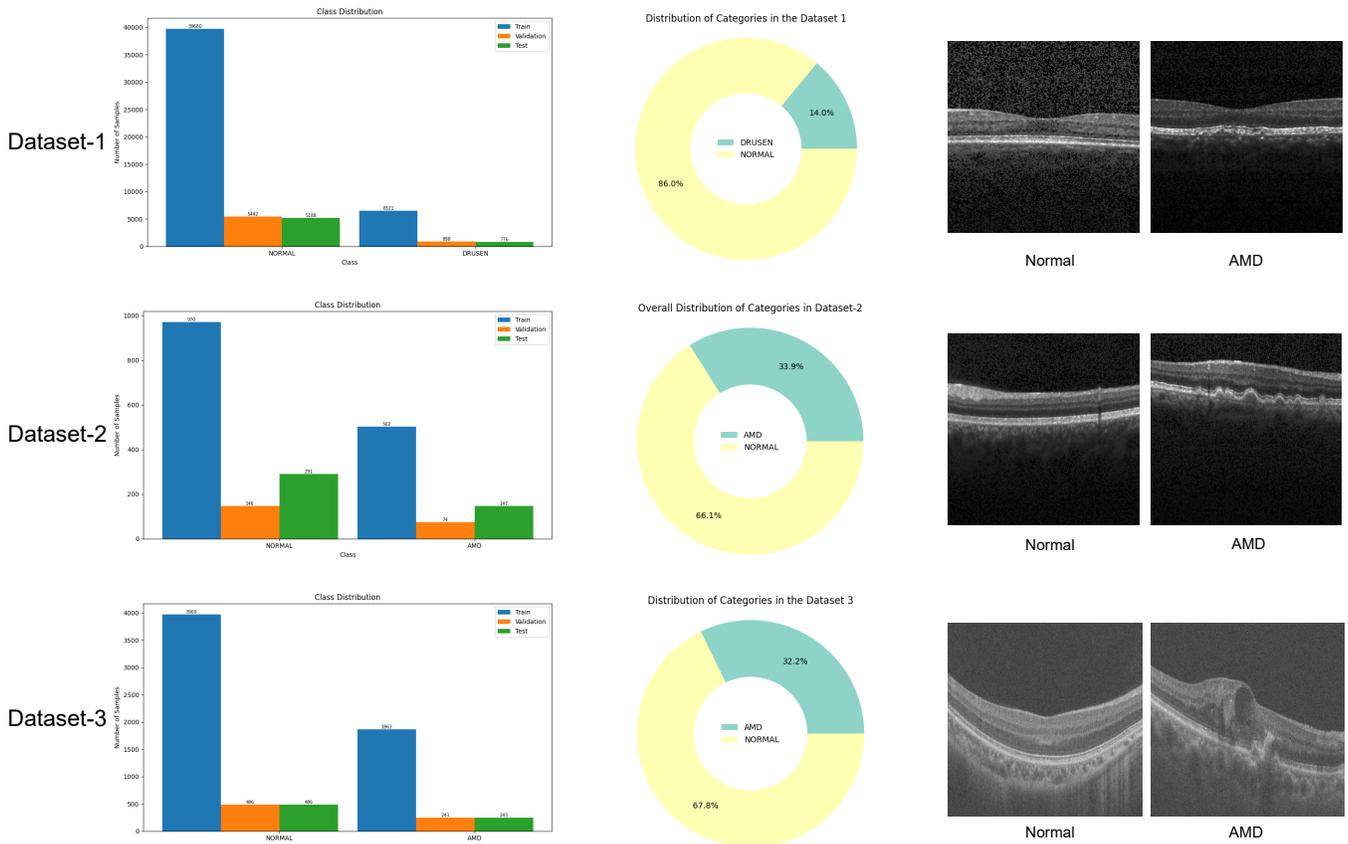}}
\caption{Illustration of Normal and AMD sample OCTs from three datasets (DS1, DS2, and DS3) along with bar chart and donut chart depicting their distribution.
}
\label{fig-dataset}
\end{figure*}
\section{Related Works}
The field of medical image classification has been profoundly impacted by the evolution of DL, starting with AlexNet\cite{NIPS2012_c399862d}. AlexNet, developed in 2012, was one of the first deep convolutional neural network (CNN) models and marked a turning point in computer vision after winning the ImageNet Challenge. This success spurred the adaptation of CNNs for medical image analysis, shifting from traditional handcrafted features to automated DL techniques \cite{kim2022transfer}. Numerous backbone models such as VGG \cite{vgg}, ResNet \cite{resnet}, and Inception \cite{inception} have been designed and then evaluated for their effectiveness in medical image classification \cite{shazia2021comparative, krishnapriya2023pre, srinivas2022deep, bressem2020comparing, yang2021detection, choi2021deep, 10.1167/tvst.9.2.35, Alam:20}. OCT image classification is a notable application of DL in ophthalmic imaging and various scientific studies have emphasized the usefulness of DL techniques in the classification of OCT images for conditions like AMD, choroidal neovascularization (CNV), and diabetic macular edema (DME). The studies, such as \cite{tsuji2020classification} and \cite{leedeep}, have demonstrated the high accuracy and effectiveness of DL in distinguishing normal OCT images from those with AMD, CNV, or DME. This underscores the potential of these techniques in automated screening and the development of computer-aided diagnostic tools.
However, DL algorithms, including CNNs, require substantial data for training. In medical imaging, where data can be scarce, transfer learning and domain adaptation techniques became crucial for leveraging knowledge from source tasks to improve target tasks \cite{ ghafoorian2017transfer, 10.1167/tvst.9.2.35}.

The evolution of transformer models in natural language processing (NLP) has opened opportunities for scalable and generalized models and sparked significant interest among researchers in computer vision. The transformer model was first introduced in 2017 \cite{vaswani2017attention} for the NLP task which leverages the self-attention mechanism to learn the contextual relationship among words within a sentence. 
The vision transformer (ViT) \cite{trans001}, distinct from CNNs, utilizes multi-head attention mechanisms to understand the contextual relationships among image pixels. This design excels in learning long-range dependencies in data, enhancing its ability to interpret spatial and temporal aspects of images. By treating images as sequences of patches, ViT effectively comprehends the overall context, a crucial factor in image classification, object detection, and pose estimation. Marking a milestone with the debut of ViT, this architecture set new benchmarks in image classification, demonstrating its vast potential in computer vision applications.
It has been extensively studied in medical imaging \cite{ayana2023vision, okolo2022ievit, alshammari2022olive, wang2022semi, kihara2022detection}. Particularly in ophthalmology, researchers in \cite{wu2021vision} proposed a transformer-based method that processes fundus images by segmenting them into distinct patches for sequential classification. This method has been shown to rival CNNs in accuracy, specificity, precision, sensitivity, and quadratic weighted kappa score, underscoring the effectiveness of attention mechanisms in diagnosing diabetic retinopathy. 
However, employing a ViT architecture requires a significant amount of computational resources, posing a challenge for communities with limited computational infrastructure \cite{islam2022recent}. Moreover, the necessity of substantial datasets for efficient training presents challenges in scenarios with limited data availability, particularly in the medical field. This domain faces additional constraints due to patient confidentiality, and the annotation of data in such contexts is notably expensive. Despite these challenges, researchers are actively working on solutions to address these issues through continuous advancements in hardware, developing efficient techniques and algorithms \cite{papa2023survey}.

SSL is a new approach to computer vision that is similar to the advancements made in NLP, specifically the development of BERT \cite{devlin2018bert}. SSL focuses on deriving meaningful information from unlabeled data. In the beginning, SSL in vision relied on contrastive learning methods, such as SimCLR \cite{chen2020simple} and MoCO \cite{he2020momentum}, which learn by contrasting positive pairs against negative pairs. However, these methods have some limitations including their reliance on large batch sizes and the need for a substantial number of negative samples.

Then SSL took a significant leap with the introduction of Masked Image Modeling (MIM). MIM is inspired by the success of BERT-like models in NLP. The masked autoencoder (MAE) \cite{he2022masked} was a pioneering approach in this domain, focusing on reconstructing masked portions of the input data. This approach allowed the model to learn robust feature representations by understanding the underlying structure of the visual data. SimMIM \cite{simmim} built on this approach and offered a simplified yet effective method by directly predicting the pixel values of the masked patches in an image. This approach marked a departure from the complexity of earlier models and demonstrated the effectiveness of direct pixel-level prediction tasks.

Further advancement in SSL for computer vision was marked by the introduction of BEiT \cite{bao2021beit}. BEiT integrated BERT-like pre-training methods into image processing. BEiT's approach to predicting masked portions of images illustrated a significant advancement in self-supervised learning paradigms, bridging the gap between language and vision modalities.

Other works underlined \cite{fang2022self, qiu2019self, jing2020self} the growing significance of SSL in ophthalmology-focused deep learning research. They demonstrate how SSL can be leveraged to overcome challenges such as the scarcity of labeled data and the need for patient-specific diagnostic tools. Moreover, they highlight the versatility of SSL in handling different data types – from 2D OCT images to complex volumetric data – making it a promising approach for future advancements in ophthalmological diagnostics and treatment planning.

These developments show the dynamic nature of SSL and how techniques from NLP can be effectively adapted to solve complex problems in computer vision. Each of these methodologies contributes to the broader understanding of how self-supervised learning can be leveraged to extract meaningful information from vast amounts of unlabeled visual data.

\section{Methodology}

In this work, we have leveraged pre-trained weights from an SSL MAE network \cite{he2022masked} with SwinV2 \cite{liu2022swin} as a backbone, demonstrating a novel approach concerning the classification of Normal Vs. AMD from OCT images by comparing their performance against the baseline model.

Our proposed framework comprises three integral stages. 

\begin{itemize}
\item[](1) Self-Supervised Pre-training: In this stage, self-supervised pre-training was conducted on unlabeled OCT images using transformer-based MAE to acquire visual representations. Following the completion of this training, the learned weights were transferred to a supervised classifier model. 
\item [] (2) Supervised Fine-tuning:
Subsequently, supervised fine-tuning was performed on labeled OCT datasets using the transferred weights to enhance the model's classification capabilities.
\item [](3) Baseline Training: We utilized ResNet50 as the baseline model to compare the performance of the proposed models. 

\end{itemize}
\subsection{Self-Supervised Pre-training}
SSL is a technique that enables a model to train itself from unlabeled data by understanding the structure or representation of the data. An example of an SSL technique is the MAE \cite{he2022masked}, which randomly masks some parts of the input data and trains the model to learn the representation of the given data to reconstruct the original input. 

The MAE trained in this study was primarily composed of two components: an encoder and a decoder. The image, resized to ($224\times $224) was fed into the encoder portion of the MAE, which then applied a patch operation ($16\times$16 patches) to randomly mask a portion (70\%) of the input image and finally processed it through a transformer encoder. In the encoder, we utilized three distinct networks—ViT \cite{trans001}, Swin\cite{liu2021swin}, and SwinV2 \cite{liu2022swin} as a backbone to conduct a comprehensive study on their performance.
\subsubsection{ViT-based MAE} 
    In ViT-based MAE, the encoder comprised an embedding dimension of 1024 and 4 attention heads, repeated for 6 layers. The encoder's final output was the set of features representing a higher-level abstraction of the original input image.

    The features were taken as input and processed by the decoder through the transformer layers. The transformer had an embedding dimension of 1024 and 4 attention heads, which were repeated for 4 layers. After passing through a linear layer to get the patches, masking was applied, and finally, the image was reconstructed.
    
\subsubsection{Swin-based MAE} 
    The Swin transformer-based MAE uses an encoder built with a Swin transformer backbone with an embedding size of 96. The number of layers in each stage of the Swin transformer architecture is (2, 2, 18, 2), which indicates the number of layers at each stage. In each stage, the model uses shifted window attention mechanisms to focus on local information within 4x4 patches. This gradually builds a global understanding by connecting windows with shifts. The attention heads are set to (6, 12, 24, 48), which doubles with each stage. This enables the model to attend to more intricate details, indicating that different layers of the encoder employ varying attention heads, capturing hierarchical features in the input image.

The decoder had an embedding size of 768, which allowed for a more expressive representation in the decoding process. The decoder network had a similar number of attention heads and layers as the encoder. It involved a patch-expanding mechanism and Swin transformer layers, which were configured to restore the spatial dimensions of the encoded features. 

The layer-wise design was built to gradually reconstruct the original image dimensions, ensuring that the decoder could effectively decode the encoded representation obtained by the encoder.

\subsubsection{SwinV2-based MAE} 
   We used a SwinV2-based MAE for this task, taking advantage of the SwinV2 network's superior performance. While we kept the Swin-based decoder, we switched the encoder for SwinV2 to address issues with training stability, high-resolution processing, and data efficiency. SwinV2's improved performance over Swin is perfect for our need for both detail and efficiency. The encoder configuration was set using an embedding dimension of 96, depths of (2, 2, 6, 2), and attention heads of (3, 6, 12, 24). This tailored approach allowed the SwinV2-based MAEs to excel at extracting intricate details.

\subsection{ Supervised Fine-Tuning}
For the classifier architecture, a classification head was added to the model in place of the decoder to use MAE as a classifier. The classification head took the features from the encoder part and ran through a linear layer to produce class logits. These class logits were then used for classification tasks. 

The linear layer was comprised of three successive dense layers each accompanied by Rectified Linear Unit (ReLU) activation functions, which progressively transform the input image into highly encoded features. The initial layer had an input size equal to the dimension of the positional embedding from the encoder part and an output size of 512. The second layer refined this feature further by mapping it to a 256-dimensional space, followed by the third layer, which compressed it to a final 128-dimensional feature vector. 

This hierarchical transformation aimed to distill and enhance the discriminative characteristics of the input, facilitating the subsequent classification task. This linear layer learned weights during training. The softmax function was then used to transform the class logits into class probabilities, enabling the model to predict the respective classes.

The approach involved fine-tuning one dataset, followed by evaluating the model's performance on the test set and two other test sets from other datasets as well to assess its robustness. This cross-evaluation procedure was repeated for all three datasets, providing a comprehensive analysis of the model's adaptability across diverse datasets. The overall algorithm of this two-phase training process is shown in \ref{alg:framework}.

Integrating self-supervised pre-training, and supervised fine-tuning in this methodology, establishes a robust framework, OCT-SelfNet for the classification of retinal diseases in OCT images. The combination of the MAE architecture and the subsequent classifier model aims to capitalize on the learned representations to improve the model's generalization to new, unseen data.

\subsection{Baseline Model}
Studies demonstrate the adaptability of the ResNet50 model in handling complex tasks like AMD detection and diabetic retinopathy classification using OCT images, showcasing its effectiveness of this architecture in medical imaging analysis \cite{leingang2023automated, SOTOUDEHPAIMA2022105368, han2022classifying, xu2023resnet}. We used it as the reference model to compare with the SSL model.  ResNet50 architecture started with a 7×7 kernel convolution and a max pooling layer, followed by a series of convolutional layers with varying sizes and numbers of kernels. Repeated in specific patterns, these layers enhanced the network's ability to extract and process complex features from images. After 50 convolutional layers, the network concluded with average pooling and fully connected layers with two nodes using softmax activation for binary classification (Normal Vs. AMD).

The network size and FLOPS of both the proposed and baseline models are detailed in Table \ref{tab:model-details}  highlighting the efficiency and compactness of the OCT-SelfNet-SwinV2 in comparison to other models.

\begin{algorithm}
\caption{An algorithm for Self-Supervised Pre-training and Supervised Fine-Tuning for OCT Classification}\label{alg:framework}
\begin{algorithmic}[1]
\Require Unlabeled OCT images $X_{\text{unlabeled}}$, Labeled OCT images $X_{\text{labeled}}$

\subsection*{\textbf{Self-Supervised Pre-training:}}
\State Initialize OCT-SelfNet, a self-supervised neural network model $M_{\text{SS}}$ with parameters $\theta_{\text{SS}}$
\State Train $M_{\text{SS}}$ on $X_{\text{unlabeled}}$ using a self-supervised objective:
  \[ \theta_{\text{SS}} = \underset{\theta_{\text{SS}}}{\arg\min} L_{\text{SS}}(\theta_{\text{SS}}, X_{\text{unlabeled}}) \]

\State Save the learned weights $\theta_{\text{SS}}$.

\subsection*{\textbf{Supervised Fine-tuning:}}

\State Initialize the classifier model, OCT-SelfNet $M_{\text{classifier}}$ with parameters $\theta_{\text{classifier}}$.
\State Load weights from $M_{\text{SS}}$ encoder to $M_{\text{classifier}}$

$M_{\text{classifier}}$:$\theta_{\text{classifier}} \leftarrow \theta_{\text{SS}}$;

\State Fine-tune $M_{\text{classifier}}$ on $X_{\text{labeled}}$ using a supervised objective:
  \[ \theta_{\text{classifier}} = \underset{\theta_{\text{classifier}}}{\arg\min} L_{\text{classifier}}(\theta_{\text{classifier}}, X_{\text{labeled}}) \]
\State Save the fine-tuned classifier weights $\theta_{\text{classifier}}$.

\subsection*{\textbf{Cross-Evaluation on Test set of different dataset:}}

\For{each test set, $X_{\text{test}}$ in dataset $D_i$ }

    \State Evaluate $M_{\text{classifier}}$ on the test set  $X_{\text{test}}$ of $D_i$.
    \State Record classification metrics: 
    
    $Metrics_i = L_{\text{classification}}(\theta_{\text{classifier}}, X_{\text{test}}^{(D_i)})$

\EndFor

\end{algorithmic}
\end{algorithm}





\begin{figure*}
\centerline{\includegraphics[width=.98\linewidth]{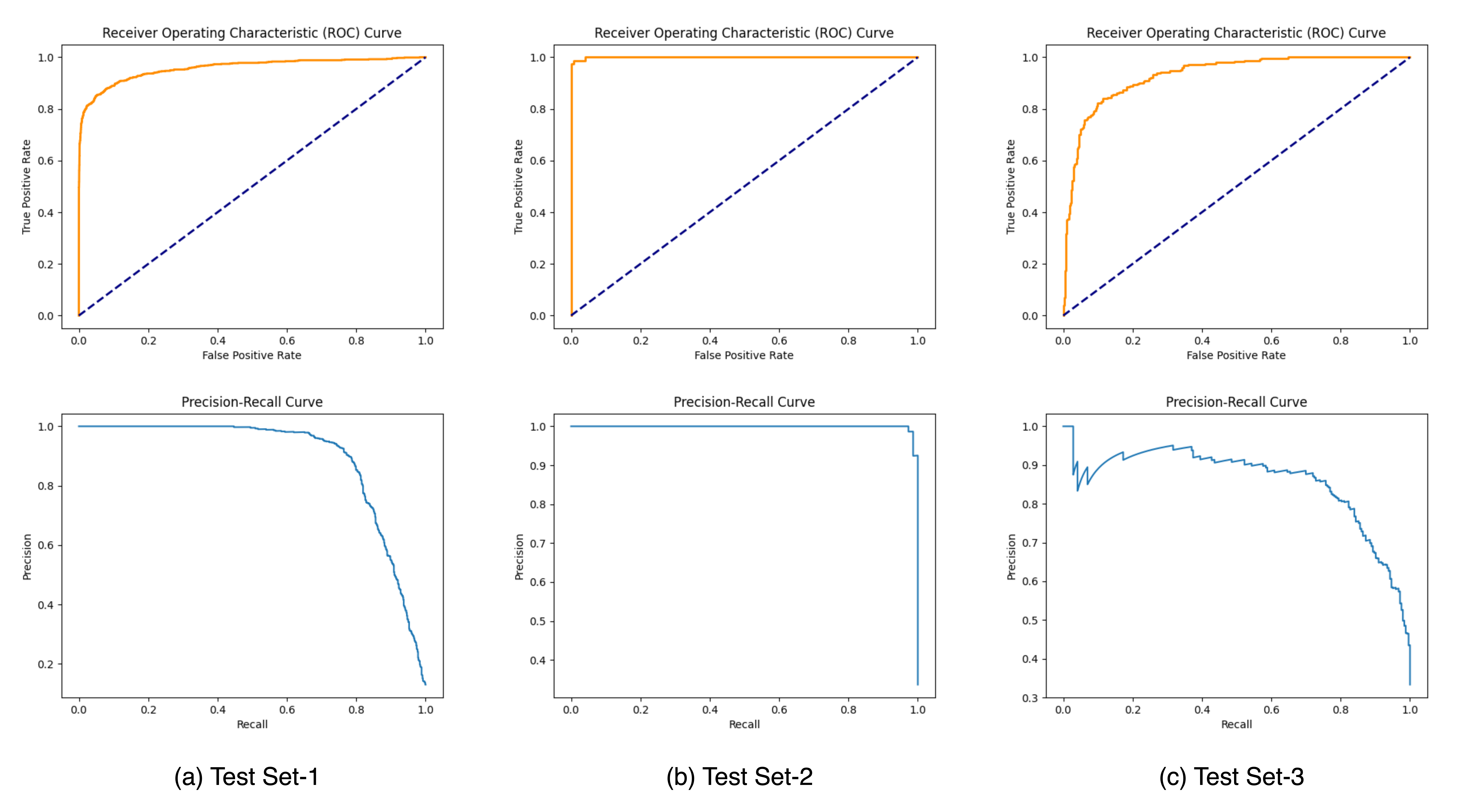}}
\caption{Evaluation of AUC-ROC and AUC-PR for Test Set-1, Test Set-2, and Test Set-3 after fine-tuning on Dataset-1 with OCT-SelfNet-Swinv2 and assessing performance on other test sets.}
\label{fig_fine_tuned_data_1_curve}
\end{figure*}

\section{Datasets}

Our experiments were carried out on three distinct datasets, DS1, DS2, and DS3 to ensure a comprehensive evaluation of the proposed methodology for binary image classification (Normal vs. AMD). The use of those different datasets enables the training of models using varied distributions and provides richer comparisons of model performance across various test sets.

\subsection{DS1}
DS1 encompasses a total of 109,559 OCT retinal (Spectralis OCT, Heidelberg Engineering, Germany) images which are classified into four categories: Normal, CNV, DME, and Drusen and put into two folders train and test. We discovered that there were identical images present in both the train and test sets as stated before in this study \cite{sinagh}. So we followed their approach to clean the data and after that, we were left with 101,565 images. Then we annotated Drusen images as AMD.

\subsection{DS2}
The DS2 dataset comprises retinal images from 45 subjects, with 15 subjects each belonging to the categories of Normal retinas, AMD, and DME. For fine-tuning purposes, we utilized data from the first ten subjects, while the rest of the data was used to test the model. All OCT volumes were obtained using Heidelberg Engineering Spectralis SD-OCT in protocols approved by the IRB \cite{srinivasan}.

\subsection{DS3}
DS3 is a dataset comprising OCT images from 500 subjects. These images were captured using two different fields of view: 3-mm and 6-mm. A single 3-mm file contains 304 scans of an individual patient, while a 6-mm file contains 400 scans. Our focus was on the slice images of the fovea since they capture the most prominent features of the retina, specifically, we analyzed image numbers 160-240 from the 6-mm scans. We considered peripheral retinal sections to have limited significance in classification. All OCT images were captured using a spectral-domain OCT system with a center wavelength of 840 nm (RTVue-XR, Optovue, CA) \cite{li2020octa}.

During the self-supervised pre-training stage, the model was trained with all classes from those three datasets, allowing it to comprehensively grasp the intricacies of representation learning. This inclusive training approach enabled the network to capture a broad spectrum of features and patterns present in the diverse classes. However, in the supervised fine-tuning stage, we focused on binary classification tasks concentrating solely on AMD and Normal categories.

A general overview of these three datasets is provided in Table \ref{tab:dataset-details} with the distribution of data across the training, validation, and test sets along with the count of normal and AMD in each set. In Figure \ref{fig-dataset}, examples of AMD and Normal sample images from the three datasets are presented, accompanied by a bar chart and donut chart illustrating their distribution.

This two-phase training approach, from comprehensive pre-training to specialized fine-tuning, strategically guides the model's representation learning and optimizes its performance for the targeted binary classification objective.

\begin{table*}[]
\centering
\caption{Distribution of data across the training, validation, and test sets for the three datasets, indicating the count of normal and AMD in each set.
}
\label{tab:dataset-details}
\begin{tabular}{|l|ll|ll|ll|l|}
\hline
\multirow{2}{*}{Dataset Name} & \multicolumn{2}{c|}{Training} & \multicolumn{2}{c|}{Validation} & \multicolumn{2}{c|}{Test} & \multirow{2}{*}{Total} \\ \cline{2-7}
          & \multicolumn{1}{l|}{Normal} & AMD  & \multicolumn{1}{l|}{Normal} & AMD & \multicolumn{1}{l|}{Normal} & AMD &        \\ \hline
Dataset-1 & \multicolumn{1}{l|}{39680}  & 6521 & \multicolumn{1}{l|}{5442}   & 898 & \multicolumn{1}{l|}{5188}   & 776 & 58,505 \\ \hline
Dataset-2 & \multicolumn{1}{l|}{970}    & 502  & \multicolumn{1}{l|}{146}    & 74  & \multicolumn{1}{l|}{291}    & 147 & 2,130  \\ \hline
Dataset-3 & \multicolumn{1}{l|}{3969}   & 1863 & \multicolumn{1}{l|}{486}    & 243 & \multicolumn{1}{l|}{486}    & 243 & 7,290  \\ \hline
\end{tabular}
\end{table*}

\section{Experiments}

\subsection{ Implementation Details}

\subsubsection{Self-Supervised Pre-training Implementation Details}
We used the following hyperparameters for this stage of training:
the learning rate is set to $1.5\times10^{-4}$ and Adam optimizer with  weight decay \cite{adamw} of $0.05$, using $\beta_1$ and $\beta_2$ values of 0.9 and 0.95, respectively. The input consists of a batch of 32 images, which were normalized. The model was run for 50 epochs, and the model with the minimum validation loss was saved for subsequent fine-tuning. The NVIDIA Tesla V100 graphical processing unit (GPU) was used for the experiments.

\subsubsection{Supervised Fine-Tuning Implementation Details}
For the fine-tuning stage, we have considered the hyperparameters as follows: 
The learning rate is set to $3\times10^{-4}$ and Adam optimizer with  weight decay \cite{adamw} of $1\times10^{-6}$, using $\beta_1$ and $\beta_2$ values of 0.9 and 0.99, respectively. The input consists of a batch of 32 images, and random resized crop, random horizontal flip, color jitter, random grayscale, and ImageNet normalization were used for data augmentation during training. The model was run for 100 epochs with early stopping criteria with patience of 10 and the model with the maximum validation accuracy was saved for testing. The NVIDIA Tesla V100 GPU was used for the experiments.

\subsubsection{Baseline Model: ResNet-50 Implementation Details}
For the baseline stage, the hyperparameters were set as follows:
 The learning rate of $3\times10^{-4}$, weight decay of $10^{-6}$, batch size of 24, and Adam optimizer with decoupled weight decay \cite{adamw}, using $\beta_1$ and $\beta_2$ values of 0.9 and 0.999, respectively. The model was trained for a maximum of 100 epochs and early stopping was applied on validation loss, with a patience of 10. During the training, the best model was saved based on the validation loss for later testing. All the baseline models were trained and evaluated on an NVIDIA GeForce RTX 3060 Ti GPU. Random rotation, horizontal flip, color jittering, Gaussian blurring, and elastic transform techniques were used for data augmentation during training.

All images were resized to 224 x 224 in training and fine-tuning, and codes were implemented with CUDA 11.2, Pytorch 1.12.1, and Python 3.10.9. 
\subsection{Evaluation Metrics}
\subsubsection{Accuracy}
Accuracy is the measure of how many correct predictions were made by the model, it is calculated by dividing the total number of correct predictions by the total number of predictions. The formula for accuracy is given by equation \ref{eq_accuracy}. 

\begin{equation}\label{eq_accuracy}
Accuracy = \frac{TP+TN}{TP+TN+FP+FN}
\end{equation}
 Here, TP = True Positives, TN = True Negatives, FP = False Positives, and FN = False Negatives.

\subsubsection{AUC-ROC}
The Area Under the Receiver Operating Characteristic curve (AUC-ROC) is a metric used to evaluate a binary classifier. In the ROC curve, the true positive rate is plotted against the false positive rate for different threshold values. The AUC-ROC is the area under the ROC curve which gives a single value that summarizes the overall performance of the model across various threshold settings. Higher AUC-ROC indicates better discrimination of positive and negative classes.

\subsubsection{AUC-PR}
Similar to AUC-ROC,  The Area Under the Precision-Recall curve (AUC-PR) is a performance metric used to evaluate a binary classifier. PR curve focuses on precision and recall by plotting precision against recall for different threshold values. The AUC-PR is the area under the Precision-Recall curve which provides a single value to denote the overall performance of the model across various threshold settings. The higher the score the better the performance.

\subsubsection{F1-Score}
F1-Score is another performance evaluation metric that takes both precision and recall into consideration which makes this metric very useful in the case of data imbalance. The F1-Score is calculated using the equation \ref{eq_f1score}.
\begin{equation}\label{eq_f1score}
F1Score = \frac{2*Precision*Recall}{Precision + Recall}
\end{equation}

The value of F1-Score ranges from 0 to 1, where a higher value indicates better performance.

\subsection{Self-Supervised Pre-training Result}
\begin{figure}
\centerline{\includegraphics[width=.99\linewidth]{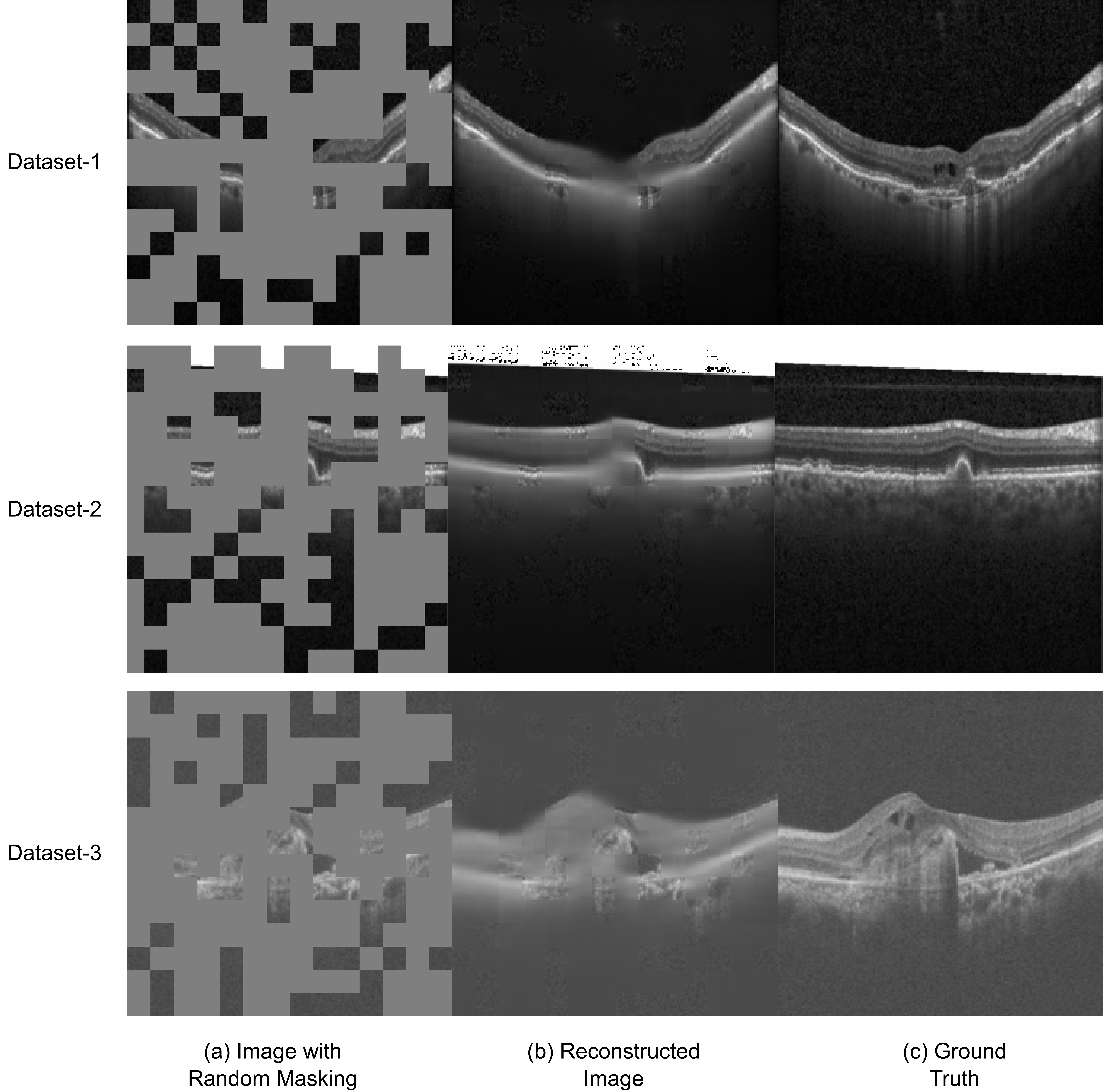}}
\caption{Qualitative visualizations of the performance of SwinV2-Based Self-Supervised MAE on Three Datasets. From left to right: Input image with randomly masked regions, reconstructed images with predicted patches, and ground truth image.}
\label{fig-mae-result}
\end{figure}

\begin{figure}
\centerline{\includegraphics[width=.98\linewidth]{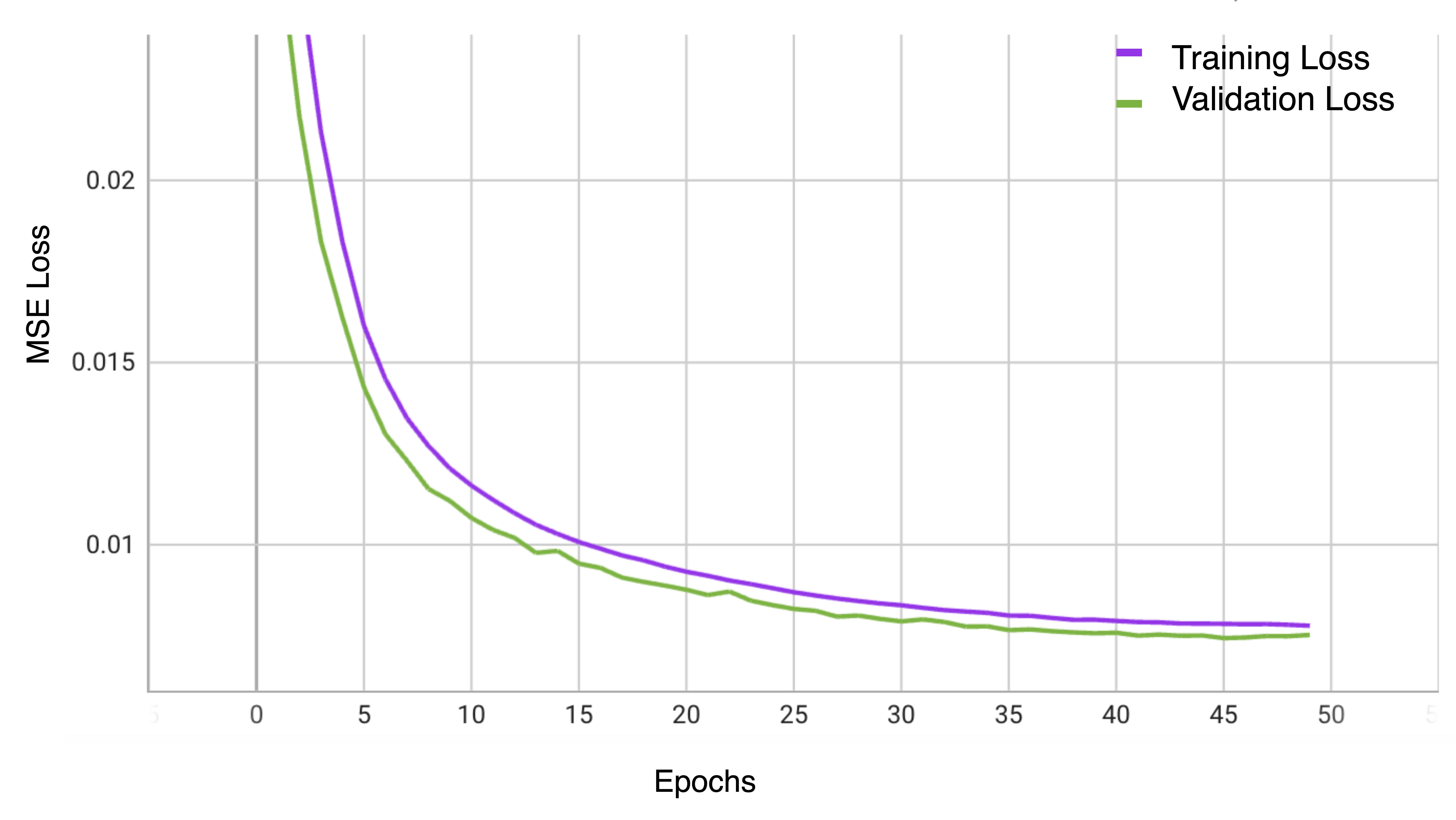}}
\caption{The training and validation MSE Loss curves of OCT-SelfNet with SwinV2 backbone.}
\label{fig_mae_loss_curve}
\end{figure}

In our evaluation, we explored the efficacy of three transformer-based networks—ViT, Swin, and SwinV2—by conducting pre-training for 50 epochs. Specifically, the SwinV2-based MAE exhibited notable proficiency, achieving a mean squared error (MSE) loss of 0.007 after 50 training epochs, as depicted in Fig \ref{fig_mae_loss_curve}. While Fig \ref{fig-mae-result} illustrates a visualization of the OCT image reconstruction from the SwinV2-based MAE, the predicted patches in the reconstructed image are not entirely clear. However, our project's primary objective was not to generate flawless reconstructions but rather to capture intricate image structures and patterns. In subsequent tasks, we utilized these pre-trained weights, leveraging their learned representations, rather than initializing the classifier network randomly.

\subsection{Supervised Fine-tuning Result}

\subsubsection{Performance Comparison with Different Encoder Network}

In our experimentation, we employed three diverse transformer-based MAE networks during the pre-training phase. In the next stage, self-supervised fine-tuning, we maintained the same encoder and leveraged transfer learning to transfer the learned weights. Additionally, a classifier network was integrated for the downstream classification tasks. 
Each supervised network underwent fine-tuning for every dataset, and evaluations were conducted across all three test sets to assess the network's performance on previously unseen test data. The performance was then compared with the baseline model ResNet-50, which underwent training on each dataset and subsequent evaluation on all three test sets. Notably, data augmentation techniques were applied throughout this experimental process.

Performance metrics, including Accuracy, AUC-ROC, AUC-PR, and F1-Score, were employed to measure the effectiveness of the different encoders. Analysis of Table \ref{tab:result-table-binary} demonstrates consistent performance of the self-supervised fine-tuning approach over the baseline model. Notably, the SwinV2-based classifier exhibits the most reliable performance across all test sets. Our OCT-SelfNet-SwinV2 model, fine-tuned on DS1, exhibited outstanding AUC-ROC scores of 0.96, 0.99, and 0.93, and AUC-PR of 0.89, 0.99, and 0.86 on Test Set-1, Test Set-2, and Test Set-3, respectively. On the other hand, the baseline model achieved AUC-ROC scores of 0.98, 0.99, and 0.56, along with AUC-PR of 0.97, 0.98, and 0.70 on the same test sets. This consistent pattern was observed across other datasets as well, affirming the reliability and versatility of our proposed methodology, with our method outperforming the baseline in AUC-ROC (0.79, 0.99, 0.86) and AUC-PR (0.42, 0.99, 0.73) on DS2, and AUC-ROC (0.75, 0.93, 0.99) and AUC-PR (0.44, 0.84, 0.99) on DS3, compared to the baseline's AUC-ROC (0.59, 0.80, 0.54) and AUC-PR (0.33, 0.86, 0.64) scores. 

\begin{figure}
\centerline{\includegraphics[width=.99\linewidth]{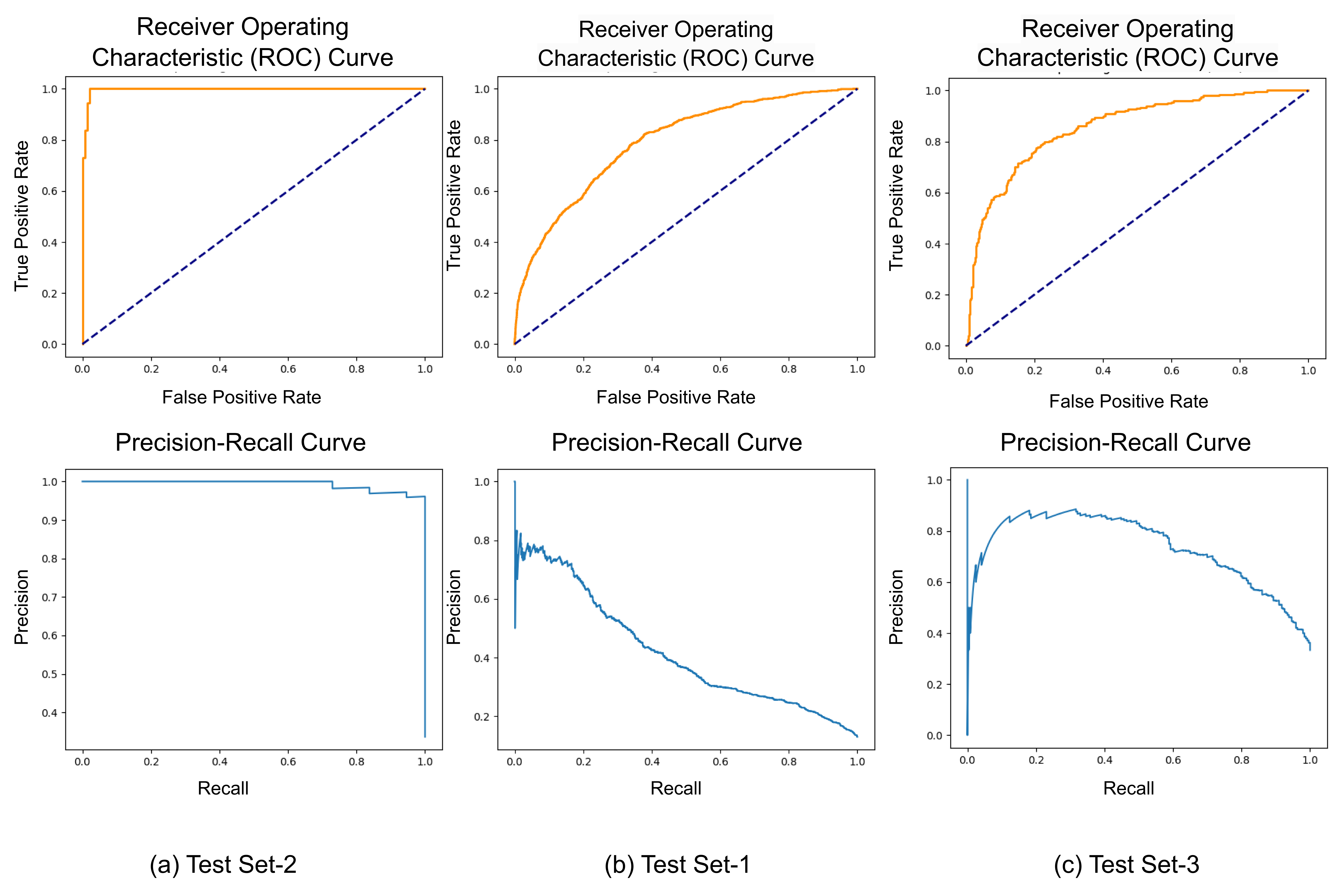}}
\caption{Evaluation of AUC-ROC and AUC-PR for Test Set-1, Test Set-2, and Test Set-3 after fine-tuning on Dataset-2 with OCT-SelfNet-SwinV2 and assessing performance on other test sets.}
\label{fig_fine_tuned_data_2_curve}
\end{figure}

\begin{figure}
\centerline{\includegraphics[width=.99\linewidth]{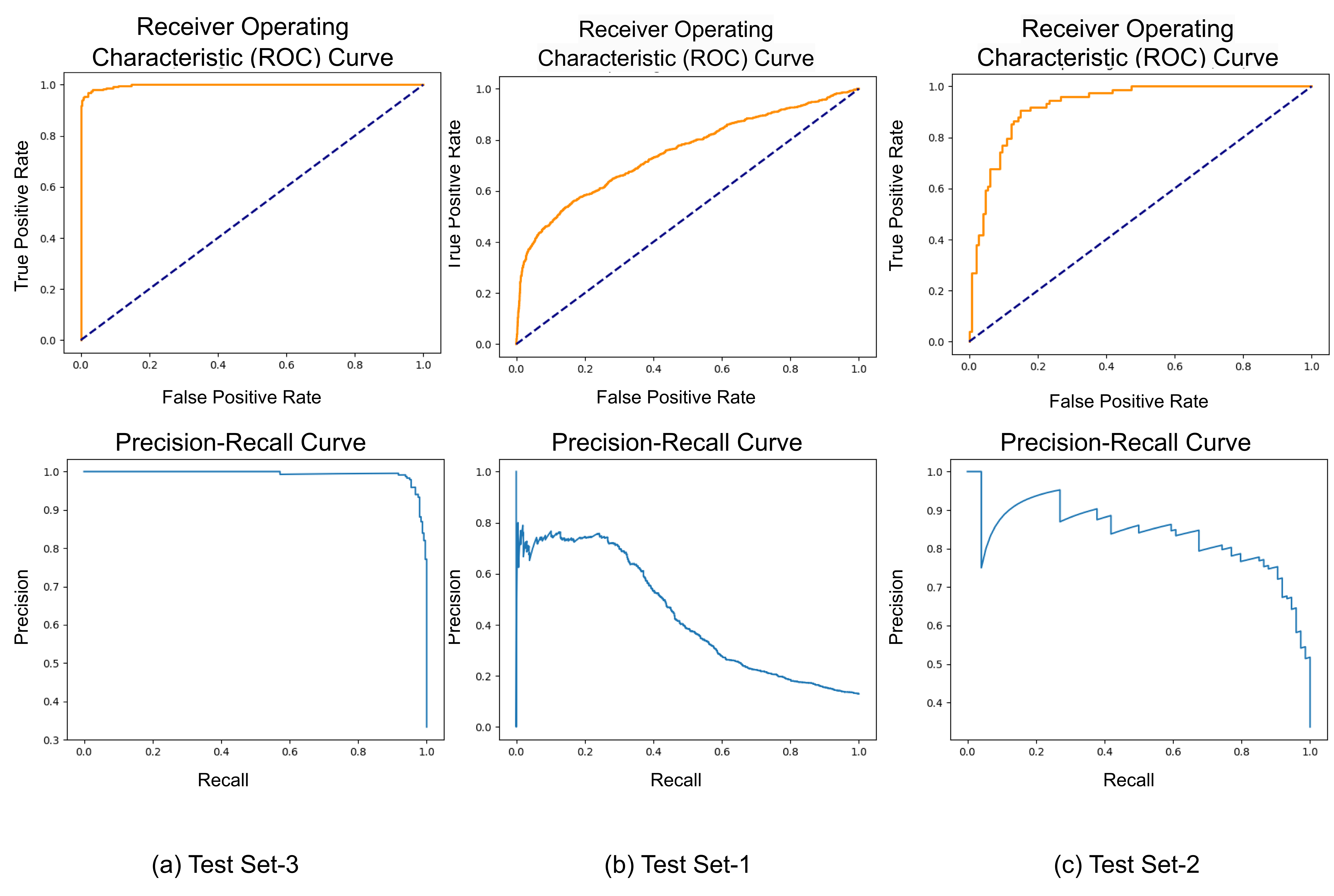}}
\caption{Evaluation of AUC-ROC and AUC-PR for Test Set-1, Test Set-2, and Test Set-3 after fine-tuning on Dataset-3 with OCT-SelfNet-SwinV2 and assessing performance on other test sets.}
\label{fig_fine_tuned_data_3_curve}
\end{figure}

When fine-tuned on DS1 with the SwinV2-based classifier and evaluated on Test Set 1, Test Set 2, and Test Set 3, the resulting AUC-ROC and AUC-PR curves are shown in Figure \ref{fig_fine_tuned_data_1_curve}, providing a visual illustration of the performance achieved through the proposed approach. Similarly, In Figure \ref{fig_fine_tuned_data_2_curve} and Figure \ref{fig_fine_tuned_data_3_curve}, the AUC-ROC and AUC-PR curves are presented for DS2 and DS3, respectively. 

In this performance analysis, our proposed model, Oct-SelfNet-SwinV2, performs better across all the datasets. Oct-SelfNet-SwinV2 achieves higher scores on all considered metrics while keeping the model size and FLOPs similar to the baseline ResNet-50 model as shown in \ref{tab:model-details}. This balance between computational cost and model effectiveness shows the model's potential for deployment in real-world applications where computational efficiency is an important factor.

\begin{table}[]
\centering
\caption{Network Details}
\label{tab:model-details}
\begin{tabular}{|l|l|l|l|}
\hline
Mode                          & Model                                    & Model Size (M) & Flops (G) \\ \hline
\multirow{3}{*}{Pre-training} & OCT-SelfNet-ViT                  &    126.88        &    10.06    \\ \cline{2-4} 
                              & OCT-SelfNet-Swinlarge            &   83.56          &    16.41   \\ \cline{2-4} 
                              & OCT-SelfNet-SwinV2               &   33.79          &  5.62
     \\ \hline
\multirow{4}{*}{Classifier}   & Resnet-50                                &    23.5       &   $4.1$    \\ \cline{2-4} 
                              & OCT-SelfNet-ViT       &       60.52 
     & 10.47       \\ \cline{2-4} 
                              & OCT-SelfNet-Swinlarge &    84.01         &    8.55    \\ \cline{2-4} 
                              & OCT-SelfNet-SwinV2    &   34.27          &   3.34     \\ \hline
\end{tabular}
\end{table}

\begin{table*}[]
\centering
\caption{Comparison of our work with the baseline methods (ResNet-50) on test sets from three datasets. The evaluation focuses on binary classification accuracy, AUC-ROC, AUC-PR, and F1-score.}
\label{tab:result-table-binary}
\begin{tabular}{|c|l|lll|lll|lll|lll|}
\hline
 &
   &
  \multicolumn{3}{c|}{Accuracy} &
  \multicolumn{3}{c|}{AUC-ROC} &
  \multicolumn{3}{c|}{AUC-PR} &
  \multicolumn{3}{c|}{F1-Score} \\ \cline{3-14} 
\multirow{-2}{*}{Train Set} &
  \multirow{-2}{*}{Classifier Name} &
  Test-1 &
  Test-2 &
  Test-3 &
  Test-1 &
  Test-2 &
  Test-3 &
  Test-1 &
  Test-2 &
  Test-3 &
  Test-1 &
  Test-2 &
  Test-3 \\ \hline
\rowcolor[HTML]{EFEFEF} 
\cellcolor[HTML]{EFEFEF} &
  Resnet50 &
  \textbf{0.99} &
  0.98 &
  0.70 &
  \textbf{0.98} &
  \textbf{0.99} &
  0.56 &
  \textbf{0.97} &
  0.98 &
  0.70 &
  \textbf{0.97} &
  0.98 &
  0.2 \\

\rowcolor[HTML]{EFEFEF} 
\multirow{-3}{*}{\cellcolor[HTML]{EFEFEF}Dataset-1} &

  OCT-SelfNet-ViT &
  0.94 &
  0.95 &
  0.78 &
  0.94 &
  \textbf{0.99} &
  0.87 &
  0.84 &
  0.97 &
  0.76 &
  0.75 &
  0.92 &
  0.56 \\ &

  OCT-SelfNet-Swinlarge &
  0.96 &
  0.97 &
  0.77 &
  0.96 &
  \textbf{0.99} &
  0.87 &
  0.90 &
  \textbf{0.99} &
  0.76 &
  0.83 &
  0.96 &
  0.51 \\ &

OCT-SelfNet-SwinV2 &
  0.96 &
  \textbf{0.99} &
  \textbf{0.88} &
  0.96 &
  \textbf{0.99} &
  \textbf{0.93} &
  0.89 &
  \textbf{0.99} &
  \textbf{0.86} &
  0.84 &
  \textbf{0.99} &
  \textbf{0.80} \\  
  \hline
 &
  Resnet50 &
  \textbf{0.77} &
  0.87 &
  0.42 &
  0.59 &
  0.80 &
  0.54 &
  0.33 &
  0.86 &
  0.64 &
  0.28 &
  0.76 &
  0.51 \\

\multirow{-3}{*}{Dataset-2} &

  OCT-SelfNet-ViT &
  0.71 &
  0.90 &
  0.54 &
  0.74 &
  0.97 &
  0.85 &
  0.30 &
  0.94 &
  0.67 &
  0.35 &
  0.87 &
  0.67 \\&

  OCT-SelfNet-Swinlarge &
  0.65 &
  0.92 &
  0.55 &
  0.75 &
  0.99 &
  0.76 &
  0.26 &
  0.96 &
  0.58 &
  0.37 &
  0.90 &
  0.57 \\&

  OCT-SelfNet-SwinV2 &
  0.72 &
  \textbf{0.99} &
  \textbf{0.55} &
  \textbf{0.79} &
  \textbf{0.99} &
  \textbf{0.86} &
  \textbf{0.42} &
  \textbf{0.99} &
  \textbf{0.73} &
  \textbf{0.39} &
  \textbf{0.98} &
  \textbf{0.59} \\
  
  \hline
\rowcolor[HTML]{EFEFEF} 
\cellcolor[HTML]{EFEFEF} &
  Resnet50 &
  0.86 &
  0.87 &
  0.95 &
  0.72 &
  0.87 &
  0.96 &
  \textbf{0.53} &
  0.84 &
  0.94 &
  \textbf{0.49} &
  0.82 &
  0.94 \\

\rowcolor[HTML]{EFEFEF} 
\multirow{-3}{*}{\cellcolor[HTML]{EFEFEF}Dataset-3} &
  OCT-SelfNet-ViT &
  \textbf{0.88} &
  0.85 &
  0.94 &
  \textbf{0.77} &
  0.90 &
  0.98 &
  0.46 &
  0.88 &
  0.97 &
  0.39 &
  0.71 &
  0.91 \\ &

    OCT-SelfNet-Swinlarge &
  0.86 &
  \textbf{0.94} &
  0.97 &
  0.75 &
  \textbf{0.95} &
  \textbf{0.99} &
 0.45 &
  \textbf{0.95} &
  \textbf{0.99} &
  0.44 &
  \textbf{0.91} &
  \textbf{0.96} \\ &

OCT-SelfNet-SwinV2 &
  \textbf{0.88} &
  0.83 &
  \textbf{0.98} &
  0.75 &
  0.93 &
  \textbf{0.99} &
  0.44 &
  0.84 &
  \textbf{0.99} &
  0.46 &
  0.72 &
  \textbf{0.96} \\ 
  
  \hline
\end{tabular}
\end{table*}

\begin{table*}[]
\centering
\caption{Comparison of SwinV2-based classifier results, with the encoder pre-trained on Train Set 1 and classifier fine-tuning on Dataset-1, followed by evaluation on Test Set 1, Test Set 2, and Test Set 3, in comparison to the baseline model.}
\label{tab:result-pretrain-oct-1}
\begin{tabular}{|l|l|lll|lll|lll|lll|}
\hline
\multirow{2}{*}{Train Set} &
  \multirow{2}{*}{Classifier Name} &
  \multicolumn{3}{c|}{Accuracy} &
  \multicolumn{3}{c|}{AUC-ROC} &
  \multicolumn{3}{c|}{AUC-PR} &
  \multicolumn{3}{c|}{F1-Score} \\ \cline{3-14} 
 &
   &
  \multicolumn{1}{l|}{Test-1} &
  \multicolumn{1}{l|}{Test-2} &
  Test-3 &
  \multicolumn{1}{l|}{Test-1} &
  \multicolumn{1}{l|}{Test-2} &
  Test-3 &
  \multicolumn{1}{l|}{Test-1} &
  \multicolumn{1}{l|}{Test-2} &
  Test-3 &
  \multicolumn{1}{l|}{Test-1} &
  \multicolumn{1}{l|}{Test-2} &
  Test-3 \\ \hline
\multirow{2}{*}{Dataset-1} &
  Resnet-50 &
  \multicolumn{1}{l|}{\textbf{0.99}} &
  \multicolumn{1}{l|}{\textbf{0.98}} &
  0.70 &
  \multicolumn{1}{l|}{\textbf{0.98}} &
  \multicolumn{1}{l|}{\textbf{0.99}} &
  0.56 &
  \multicolumn{1}{l|}{\textbf{0.97}} &
  \multicolumn{1}{l|}{0.98} &
  0.70 &
  \multicolumn{1}{l|}{\textbf{0.97}} &
  \multicolumn{1}{l|}{\textbf{0.98}} &
  0.20 \\ \cline{2-14} 
 &
  OCT-SelfNet-SwinV2 &
  \multicolumn{1}{l|}{0.96} &
  \multicolumn{1}{l|}{0.97} &
  \textbf{0.82} &
  \multicolumn{1}{l|}{0.96} &
  \multicolumn{1}{l|}{\textbf{0.99}} &
  \textbf{0.91} &
  \multicolumn{1}{l|}{0.91} &
  \multicolumn{1}{l|}{\textbf{0.99}} &
  \textbf{0.84} &
  \multicolumn{1}{l|}{0.84} &
  \multicolumn{1}{l|}{0.96} &
  \textbf{0.66} \\ \hline
\end{tabular}
\end{table*}

\begin{table*}[]
\centering
\caption{Comparison of our work with the baseline methods (ResNet-50) on test sets from three datasets, using only 50\% of the training data in Finetuning. The evaluation focuses on binary classification accuracy, AUC-ROC, AUC-PR, and F1-score.}
\label{tab:result-table-binary-half}
\begin{tabular}{|c|l|lll|lll|lll|lll|}
\hline
 &
   &
  \multicolumn{3}{c|}{Accuracy} &
  \multicolumn{3}{c|}{AUC-ROC} &
  \multicolumn{3}{c|}{AUC-PR} &
  \multicolumn{3}{c|}{F1-Score} \\ \cline{3-14} 
\multirow{-2}{*}{Train Set} &
  \multirow{-2}{*}{Classifier Name} &
  Test-1 &
  Test-2 &
  Test-3 &
  Test-1 &
  Test-2 &
  Test-3 &
  Test-1 &
  Test-2 &
  Test-3 &
  Test-1 &
  Test-2 &
  Test-3 \\ \hline
\rowcolor[HTML]{EFEFEF} 
\cellcolor[HTML]{EFEFEF} &
  Resnet50 &
  \textbf{0.99} &
  \textbf{0.99} &
  0.70 &
  \textbf{0.98} &
  \textbf{0.99} &
  0.55 &
  \textbf{0.97} &
  0.98 &
  0.70 &
  \textbf{0.97} &
  \textbf{0.98} &
  0.17 \\

\rowcolor[HTML]{EFEFEF} 
\multirow{-3}{*}{\cellcolor[HTML]{EFEFEF}Dataset-1} &

  OCT-SelfNet-ViT &
  0.94 &
  0.94 &
  0.76 &
  0.94 &
  0.98 &
  0.85 &
  0.84 &
  0.97 &
  0.79 &
  0.74 &
  0.91 &
  0.46 \\

&
  OCT-SelfNet-Swinlarge &
  0.95 &
  0.97 &
  \textbf{0.78} &
  0.94 &
  \textbf{0.99} &
  0.86 &
  0.85 &
  \textbf{0.99} &
  0.75 &
  0.77 &
  0.96 &
  \textbf{0.61} \\

 & 
OCT-SelfNet-SwinV2 &
  0.96 &
  0.97 &
  \textbf{0.78} &
  0.96 &
  \textbf{0.99} &
  \textbf{0.93} &
  0.89 &
  \textbf{0.99} &
  \textbf{0.83} &
  0.82 &
  0.95 &
  0.54 \\

  \hline
 &
  Resnet50 &
  \textbf{0.79} &
  0.85 &
  0.33 &
  0.56 &
  0.78 &
  0.50 &
  0.28 &
  0.85 &
  0.66 &
  0.23 &
  0.71 &
  0.50 \\

\multirow{-3}{*}{Dataset-2} &

  OCT-SelfNet-ViT &
  0.70 &
  0.87 &
  \textbf{0.49} &
  0.73 &
  0.96 &
  \textbf{0.83} &
  0.28 &
  0.94 &
  \textbf{0.67} &
  0.35 &
  0.83 &
 \textbf{0.57} \\

  &

  OCT-SelfNet-Swinlarge &
  0.68 &
  0.93 &
  0.47 &
  0.70 &
  \textbf{0.99} &
  0.71 &
  0.26 &
  0.96 &
  0.55 &
  0.32 &
  \textbf{0.91} &
  0.55 \\

  &

  OCT-SelfNet-SwinV2 &
  0.72 &
  \textbf{0.94} &
  0.34 &
  \textbf{0.76} &
  \textbf{0.99} &
  0.63 &
  \textbf{0.31} &
  \textbf{0.98} &
  0.41 &
  \textbf{0.37} &
  \textbf{0.91} &
  0.50 \\

  \hline
\rowcolor[HTML]{EFEFEF} 
\cellcolor[HTML]{EFEFEF} &
  Resnet50 &
  0.77 &
  \textbf{0.87} &
  \textbf{0.98} &
  \textbf{0.75} &
  0.89 &
  \textbf{0.99} &
  \textbf{0.54} &
  0.85 &
  \textbf{0.98} &
  \textbf{0.45} &
  \textbf{0.83} &
  \textbf{0.98} \\

\rowcolor[HTML]{EFEFEF} 
\multirow{-3}{*}{\cellcolor[HTML]{EFEFEF}Dataset-3} &
  OCT-SelfNet-ViT &
  \textbf{0.88} &
  0.79 &
  0.95 &
  \textbf{0.75} &
  0.92 &
  \textbf{0.99} &
  0.41 &
  0.86 &
  \textbf{0.98} &
  0.32 &
  0.58 &
  0.92 \\

 &

    OCT-SelfNet-Swinlarge &
  \textbf{0.88} &
  0.78 &
  0.95 &
  0.72 &
  0.87 &
  \textbf{0.99} &
  0.41 &
  0.79 &
  \textbf{0.98} &
  0.28 &
  0.55 &
  0.92 \\

  &

OCT-SelfNet-SwinV2 &
  \textbf{0.88} &
  0.79 &
  0.95 &
  0.64 &
  \textbf{0.96} &
  0.98 &
  0.33 &
  \textbf{0.93} &
  \textbf{0.98} &
  0.29 &
  0.55 &
  0.92 \\

  \hline
\end{tabular}
\end{table*}

\begin{table*}[]
\centering
\caption{With No Augmentation, Comparison of Our Work with Baseline Methods (ResNet-50) on Test Sets from Three Datasets in Terms of Binary Classification Accuracy, AUC-ROC, AUC-PR, and F1-Score.}
\label{tab:result-table-binary-no-aug}
\begin{tabular}{|c|l|lll|lll|lll|lll|}
\hline
 &
   &
  \multicolumn{3}{c|}{Accuracy} &
  \multicolumn{3}{c|}{AUC-ROC} &
  \multicolumn{3}{c|}{AUC-PR} &
  \multicolumn{3}{c|}{F1-Score} \\ \cline{3-14} 
\multirow{-2}{*}{Train Set} &
  \multirow{-2}{*}{Classifier Name} &
  Test-1 &
  Test-2 &
  Test-3 &
  Test-1 &
  Test-2 &
  Test-3 &
  Test-1 &
  Test-2 &
  Test-3 &
  Test-1 &
  Test-2 &
  Test-3 \\ \hline
\rowcolor[HTML]{EFEFEF} 
\cellcolor[HTML]{EFEFEF} &
  Resnet50 &
  \textbf{0.99} &
  \textbf{0.97} &
  0.67 &
  \textbf{0.99} &
  0.97 &
  0.50 &
  \textbf{0.98} &
  0.95 &
  0.67 &
  \textbf{0.97} &
  0.95 &
  0 \\

\rowcolor[HTML]{EFEFEF} 
\multirow{-3}{*}{\cellcolor[HTML]{EFEFEF}Dataset-1} &

  OCT-SelfNet-ViT &
  0.95 &
  \textbf{0.97} &
  \textbf{0.79} &
  0.94 &
  \textbf{0.99} &
  \textbf{0.87} &
  0.85 &
  \textbf{0.99} &
  \textbf{0.83} &
  0.78 &
  0.95 &
  \textbf{0.54} \\

&
  OCT-SelfNet-Swinlarge &
  0.96 &
  0.97 &
  0.66 &
  0.97 &
  0.99 &
  0.66 &
  0.90 &
  0.99 &
  0.45 &
  0.81 &
  0.96 &
  0.17 \\

 & 
OCT-SelfNet-SwinV2 &
  0.96 &
  0.92 &
  0.66 &
  0.95 &
  0.97 &
  0.48 &
  0.88 &
  0.96 &
  0.36 &
  0.81 &
  0.88 &
  0.18 \\

  \hline
 &
  Resnet50 &
  \textbf{0.69} &
  0.91 &
  \textbf{0.33} &
  \textbf{0.60} &
  0.88 &
  \textbf{0.50} &
  \textbf{0.37} &
  0.91 &
  \textbf{0.67} &
  \textbf{0.29} &
  0.85 &
  \textbf{0.50} \\

\multirow{-3}{*}{Dataset-2} &

  OCT-SelfNet-ViT &
  0.42 &
  \textbf{0.95} &
  \textbf{0.33} &
  0.59 &
  \textbf{0.99} &
  0.10 &
  0.15 &
  \textbf{0.99} &
  0.20 &
  0.25 &
  \textbf{0.94} &
  \textbf{0.50} \\

  &

  OCT-SelfNet-Swinlarge &
  0.49 &
  0.94 &
  \textbf{0.33} &
  0.57 &
  \textbf{0.99} &
  0.46 &
  0.15 &
  0.97 &
  0.31 &
  0.25 &
  0.92 &
  \textbf{0.50} \\

  &

  OCT-SelfNet-SwinV2 &
  0.54 &
  \textbf{0.96} &
  \textbf{0.33} &
  0.55 &
  0.97 &
  0.18 &
  0.14 &
  0.85 &
  0.21 &
  0.26 &
  \textbf{0.95} &
  \textbf{0.50} \\

  \hline
\rowcolor[HTML]{EFEFEF} 
\cellcolor[HTML]{EFEFEF} &
  Resnet50 &
  \textbf{0.87} &
  0.66 &
  0.92 &
  0.50 &
  0.50 &
  0.94 &
  \textbf{0.57} &
  0.67 &
  0.90 &
  0 &
  0 &
  0.89 \\

\rowcolor[HTML]{EFEFEF} 
\multirow{-3}{*}{\cellcolor[HTML]{EFEFEF}Dataset-3} &
  OCT-SelfNet-ViT &
  \textbf{0.87} &
  0.66 &
  0.90 &
  0.46 &
  0.71 &
  0.90 &
  0.13 &
  0.65 &
  \textbf{1} &
  0 &
  0 &
  0.70 \\
 &

OCT-SelfNet-Swinlarge &
  0.74 &
  0.63 &
  \textbf{0.98} &
  \textbf{0.56} &
  0.38 &
  \textbf{0.99} &
  0.17 &
  0.27 &
  \textbf{0.99} &
  \textbf{0.19} &
  0.18 &
  \textbf{0.97} \\

  &

OCT-SelfNet-SwinV2 &
  0.67 &
  \textbf{0.87} &
  \textbf{0.98} &
  0.52 &
  \textbf{0.86} &
  \textbf{0.99} &
  0.14 &
  \textbf{0.76} &
  0.94 &
  0.18 &
  \textbf{0.80} &
  \textbf{0.97} \\

  \hline
\end{tabular}
\end{table*}

\subsubsection{Performance Comparison in Limited Data Settings}
In this ablation study, the training set was intentionally reduced to 50\%, and subsequent fine-tuning was conducted for each dataset with the augmentation techniques. The objective was to assess the performance of the OCT-SelfNet classifier in comparison to the baseline model under insufficient data conditions.

Examination of Table \ref{tab:result-table-binary-half} indicates that, even with a reduction in training data, the proposed method consistently outperforms the baseline model. Specifically, the discrepancy among the scores of unseen test sets compared to the fine-tuned dataset's test set is considerable for the baseline model. In contrast, the self-supervised fine-tuning approach exhibits minimal gaps, showcasing better generalization capabilities. When fine-tuned on the reduced DS1 training set, the OCT-SelfNet-SwinV2 model exhibited AUC-ROC scores of 0.96, 0.99, 0.93, and AUC-PR of 0.89, 0.99, 0.83 on Test Set-1, Test Set-2, and Test Set-3, respectively. In comparison, the baseline model achieved AUC-ROC scores of 0.98, 0.99, and 0.55, accompanied by AUC-PR of 0.97, 0.98, and 0.70 on the same test sets. This consistent pattern extended to other datasets, as summarized in Table \ref{tab:result-table-binary-half}.
This finding underscores the robustness and improved adaptability of the proposed methodology, particularly in scenarios with limited training data.

\subsubsection{Performance Evaluation on Unseen Datasets Settings}
In this experiment, the OCT-SelfNet-SwinV2 underwent pre-training exclusively with DS1, and the acquired weights were subsequently transferred to the classifier network. The classifier network was then fine-tuned on DS1, and evaluations were conducted on Test Set 1, Test Set 2, and Test Set 3.
For effective pre-training and optimal learning representation, a larger dataset is essential, therefore, DS1 was selected as it has the highest number of samples among the three datasets. The objective of this experiment was to assess the performance of our proposed method on previously unseen datasets. 

The results of this experiment, presented in Table \ref{tab:result-pretrain-oct-1}, demonstrate that the proposed method consistently outperforms or matches the baseline model across diverse test sets, demonstrating AUC-ROC scores of 0.96, 0.99, 0.91, and AUC-PR scores of 0.91, 0.99, 0.84. In comparison, the baseline model achieves AUC-ROC of 0.98, 0.99, 0.56, and AUC-PR of 0.97, 0.98, 0.70.  
 Notably, the minimal score gaps observed in OCT-SelfNet-SwinV2 for diverse test sets emphasize the generalization capabilities of our proposed approach, which further affirms its efficacy in handling a wide range of datasets.

\subsubsection{Evaluating the Influence of Augmentation Techniques on the Performance}
In this experiment, all three self-supervised fine-tuned networks and the baseline model were evaluated without the application of augmentation techniques. Table \ref{tab:result-table-binary-no-aug} displays the scores obtained from this experiment, highlighting the significant impact of augmentation methods. The results indicate a noticeable decrease in performance for both our proposed method and the baseline model in the absence of augmentation. This underscores the crucial role played by augmentation techniques in enhancing the performance of the models, emphasizing their positive influence on the overall effectiveness of the proposed approach.

\section{Conclusion}
In this article, we have investigated the effectiveness of a self-supervised DL framework utilizing the SwinV2 backbone for the detection of eye diseases through OCT images. While the shifted window-based transformer model has gained attention in research, its application in OCT analysis has been relatively unexplored. This article presents a comprehensive two-phase framework that adeptly learns feature representations during the pre-training stage from a multi-modal dataset and leverages these learned weights for the downstream target task. Our experiments, conducted across three diverse datasets with cross-evaluation and an extensive ablation study, yield promising results. Our proposed framework OCT-SelfNet-SwinV2 consistently outperforms the baseline, showcasing superior AUC-ROC scores across DS1, DS2, and DS3. Fine-tuned on DS1, it excels with scores of 0.96, 0.99, and 0.93 on Test Set-1, Test Set-2, and Test Set-3, surpassing baseline scores of 0.98, 0.99, and 0.56. DS2 achieves 0.79, 0.99, and 0.86 against the baseline's 0.59, 0.80, and 0.54. DS3 exhibits scores of 0.75, 0.93, and 0.99, surpassing the baseline's 0.72, 0.87, and 0.96. These improvements highlight our model's consistency and the potential of our approach to enhance the robustness and generalization of eye disease detection in real-world clinical scenarios. 


{\small
\bibliographystyle{ieee_fullname}
\bibliography{egbib}
}

\end{document}